\documentclass[final]{cvpr}

\usepackage{times}
\usepackage{epsfig}
\usepackage{graphicx}
\usepackage{amsmath}
\usepackage{amssymb}
\usepackage{multirow}
\usepackage[noend]{algpseudocode}
\usepackage{algorithmicx,algorithm}

\usepackage[pagebackref=true,breaklinks=true,colorlinks,bookmarks=false]{hyperref}

\def\cvprPaperID{5021} 
\def\confYear{CVPR 2021}

\begin{document}

\title{OTA: Optimal Transport Assignment for Object Detection}

\author{Zheng Ge$^{1,2}$,~~~ Songtao Liu$^2$\thanks{Corresponding author},~~~ Zeming Li$^2$,~~~ Osamu Yoshie$^1$,~~~  Jian Sun$^2$\\
$^1$Waseda University, $^2$Megvii Technology\\
\tt\small jokerzz@fuji.waseda.jp;liusongtao@megvii.com;lizeming@megvii.com;\\
\tt\small yoshie@waseda.jp;sunjian@megvii.com}

\maketitle

\begin{abstract}

Recent advances in label assignment in object detection mainly seek to independently define positive/negative training samples for each ground-truth (\textit{gt}) object. In this paper, we innovatively revisit the label assignment from a global perspective and propose to formulate the assigning procedure as an Optimal Transport (OT) problem -- a well-studied topic in Optimization Theory. Concretely, we define the unit transportation cost between each demander (anchor) and supplier (gt) pair as the weighted summation of their classification and regression losses. After formulation, finding the best assignment solution is converted to solve the optimal transport plan at minimal transportation costs, which can be solved via Sinkhorn-Knopp Iteration. On COCO, a single FCOS-ResNet-50 detector equipped with Optimal Transport Assignment (OTA) can reach 40.7\% mAP under 1$\times$ scheduler, outperforming all other existing assigning methods. Extensive experiments conducted on COCO and CrowdHuman further validate the effectiveness of our proposed OTA, especially its superiority in crowd scenarios. The code is available at \url{https://github.com/Megvii-BaseDetection/OTA}.
\end{abstract}

\section{Introduction}\label{intro}
Current CNN-based object detectors~\cite{ssd,yolov1,focalloss,atss,fasterrcnn, borderdet, dynhead} perform a dense prediction manner by predicting the classification (\emph{cls}) labels and regression (\emph{reg}) offsets for a set of pre-defined anchors\footnote{For anchor-free detectors like FCOS~\cite{fcos}, the feature points can be viewed as shrunk anchor boxes. Hence in this paper, we collectively refer to anchor box and anchor point as ``anchor''.}. To train the detector, defining \emph{cls} and \emph{reg} targets for each anchor is a necessary procedure, which is called \emph{label assignment} in object detection.

Classical label assigning strategies commonly adopt pre-defined rules to match the ground-truth (\emph{gt}) object or background for each anchor. For example, RetinaNet~\cite{focalloss} adopts Intersection-over-Union (IoU) as its thresholding criterion for \emph{pos}/\emph{neg} anchors division. Anchor-free detectors like FCOS~\cite{fcos} treat the anchors within the center/bbox region of any \emph{gt} object as the corresponding positives. Such static strategies ignore a fact that for objects with different sizes, shapes or occlusion condition, the appropriate positive/negative (\emph{pos/neg}) division boundaries may vary.

Motivated by this, many dynamic assignment strategies have been proposed. ATSS~\cite{atss} proposes to set the division boundary for each $gt$ based on statistical characteristics. Other recent advances~\cite{freeanchor, noisyanchors, autoassign,paa} suggest that the predicted confidence scores of each anchor could be a proper indicator to design dynamic assigning strategies, \emph{i.e.}, high confidence anchors can be easily learned by the networks and thus be assigned to the related \emph{gt}, while anchors with uncertain predictions should be considered as negatives. Those strategies enable the detector to dynamically choose positive anchors for each individual \emph{gt} object and achieve state-of-the-art performance.

\begin{figure}[!t]\label{figintro}
\includegraphics[width=8.3cm]{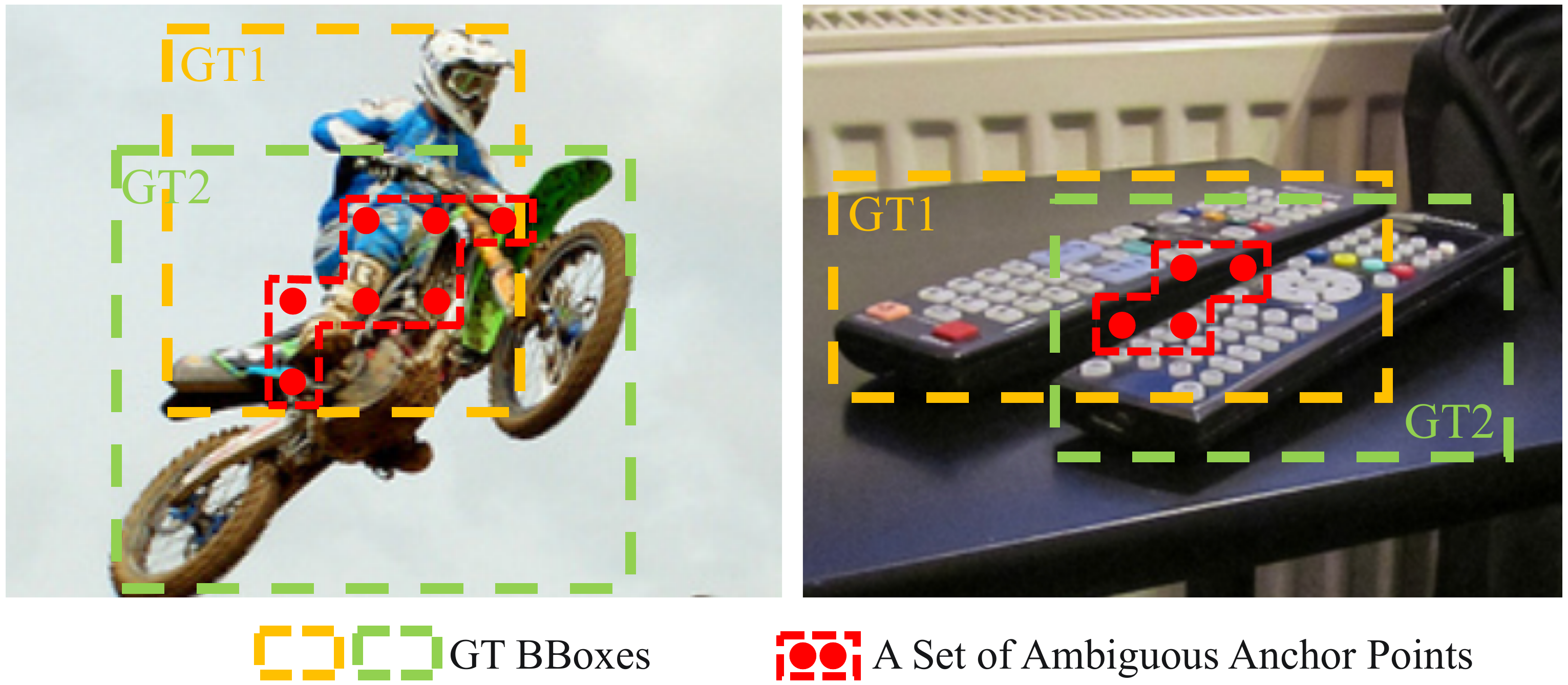}
\caption{An illustration of ambiguous anchor points in object detection. Red dots show some of the ambiguous anchors in two sample images. Currently, the assignment of these ambiguous anchors is heavily based on hand-crafted rules.}
\label{intro}
\end{figure}

However, independently assigning \emph{pos}/\emph{neg} samples for each \emph{gt} without context could be sub-optimal, just like the lack of context may lead to improper prediction. When dealing with ambiguous anchors (\emph{i.e.}, anchors that are qualified as positive samples for multiple $gt$s simultaneously as seen in Fig.~\ref{figintro}.), existing assignment strategies are heavily based on hand-crafted rules (\emph{e.g.}, Min Area~\cite{fcos}, Max IoU~\cite{paa,focalloss,atss}.). We argue that assigning ambiguous anchors to any $gt$ (or \emph{background}) may introduce harmful gradients \emph{w.r.t.} other $gt$s. Hence the assignment for ambiguous anchors is non-trivial and requires further information beyond the local view. Thus a better assigning strategy should get rid of the convention of pursuing optimal assignment for each \emph{gt} independently and turn to the ideology of global optimum, in other words, finding the global high confidence assignment for all \emph{gt}s in an image.

DeTR~\cite{detr} is the first work that attempts to consider \emph{label assignment} from global view. It replaces the detection head with transformer layers~\cite{transformer} and considers one-to-one assignment using the Hungarian algorithm that matches only one query for each \emph{gt} with global minimum loss. However, for the CNN based detectors, as the networks often produce correlated scores to the neighboring regions around the object, each \emph{gt} is assigned to many anchors (\emph{i.e.}, one-to-many), which also benefits to training efficiency. In this one-to-many manner, it remains intact to assign labels with a global view.  

To achieve the global optimal assigning result under the one-to-many situation, we propose to formulate \emph{label assignment} as an Optimal Transport (OT) problem -- a special form of Linear Programming (LP) in Optimization Theory. Specifically, we define each \emph{gt} as a supplier who supplies a certain number of labels, and define each anchor as a demander who needs one unit label. If an anchor receives sufficient amount of positive label from a certain \emph{gt}, this anchor becomes one positive anchor for that \emph{gt} . In this context, the number of positive labels each \emph{gt} supplies can be interpreted as ``how many positive anchors that \emph{gt} needs for better convergence during the training process''. The unit transportation cost between each anchor-\emph{gt} pair is defined as the weighted summation of their pair-wise \emph{cls} and \emph{reg} losses. Furthermore, as being negative should also be considered for each anchor, we introduce another supplier -- \emph{background} who supplies negative labels to make up the rest of labels in need. The cost between \emph{background} and a certain anchor is defined as their pair-wise classification loss only. After formulation, finding the best assignment solution is converted to solve the optimal transport plan, which can be quickly and efficiently solved by the off-the-shelf Sinkhorn-Knopp Iteration~\cite{sinkhorn}. We name such an assigning strategy as  Optimal Transport Assignment (OTA).

Comprehensive experiments are carried out on MS COCO~\cite{coco} benchmark, and significant improvements from OTA demonstrate its advantage. OTA also achieves the SOTA performance among one-stage detectors on a crowded pedestrian detection dataset named CrowdHuman~\cite{crowdhuman}, showing OTA's generalization ability on different detection benchmarks.


\section{Related Work}

\subsection{Fixed Label Assignment}

Determining which $gt$ (or \emph{background}) should each anchor been assigned to is a necessary procedure before training object detectors. Anchor-based detectors usually adopt IoU at a certain threshold as the assigning criterion. For example, RPN in Faster R-CNN~\cite{fasterrcnn} uses 0.7 and 0.3 as the positive and negative thresholds, respectively. When training the R-CNN module, the IoU threshold for \emph{pos}/\emph{neg} division is changed to 0.5. IoU based \emph{label assignment} is proved effective and soon been adopted by many Faster R-CNN's variants like~\cite{cascadercnn,maskrcnn,fpn,mpsr,pat,crossDA,treev2}, as well as many one-stage detectors like~\cite{yolo9000,yolov3,asff,ssd,rfb,focalloss}.

Recently, anchor-free detectors have drawn much attention because of their concision and high computational efficiency. Without anchor box, FCOS~\cite{fcos}, Foveabox~\cite{foveabox} and their precursors~\cite{yolov1,densebox,unitbox} directly assign anchor points around the center of objects as positive samples, showing promising detection performance. Another stream of anchor-free detectors~\cite{cornernet,centernet,objectsaspoints,reppointsv1,reppointsv2} view each object as a single or a set of key-points. They share distinct characteristics from other detectors, hence will not be further discussed in our paper.

Although detectors mentioned above are different in many aspects, as for \emph{label assignment}, they all adopt a single fixed assigning criterion (\emph{e.g.}, a fixed region of the center area or IoU threshold) for objects of various sizes, shapes, and categories, etc, which may lead to sub-optimal assigning results.

\begin{figure*}[!t]
\begin{center}
\includegraphics[width=16.5cm]{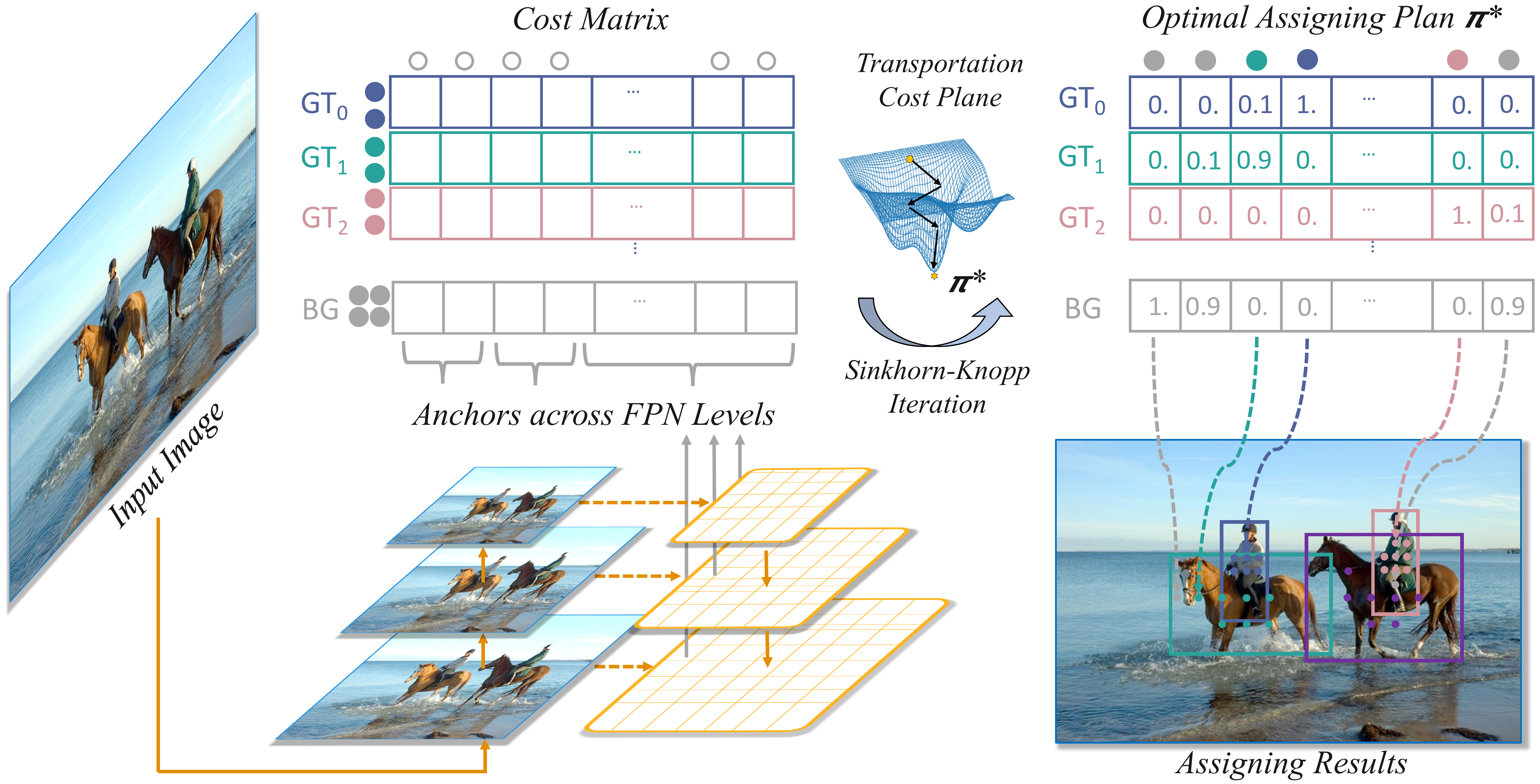}
\end{center}
   \caption{An illustration of Optimal Transport Assignment. \emph{Cost Matrix} is composed of the pair-wise \emph{cls} and \emph{reg} losses between each anchor-\emph{gt} pair. The goal of finding the best label assigning is converted to solve the best transporting plan which transports the labels from suppliers (\emph{i.e.} GT and BG) to demanders (\emph{i.e.} anchors) at a minimal transportation cost via Sinkhorn-Knopp Iteration.}
\label{model}
\end{figure*}

\subsection{Dynamic Label Assignment}

Many recent works try to make the label assigning procedure more adaptive, aiming to further improve the detection performance. Instead of using pre-defined anchors, GuidedAnchoring~\cite{guidedanchor} generates anchors based on an anchor-free mechanism to better fit the distribution of various objects. MetaAnchor~\cite{metaanchor} proposes an anchor generation function to learn dynamic anchors from the arbitrary customized prior boxes. NoisyAnchors~\cite{noisyanchors} proposes soft-label and anchor re-weighting mechanisms based on classification and localization losses. FreeAnchor~\cite{freeanchor} constructs top-k anchor candidates for each $gt$ based on IoU and then proposes a detection-customized likelihood to perform \emph{pos}/\emph{neg} division within each candidate set. ATSS~\cite{atss} proposes an adaptive sample selection strategy that adopts \emph{mean+std} of IoU values from a set of closest anchors for each $gt$ as a \emph{pos}/\emph{neg} threshold. PAA~\cite{paa} assumes that the distribution of joint loss for positive and negative samples follows the Gaussian distribution. Hence it uses GMM to fit the distribution of positive and negative samples, and then use the center of positive sample distribution as the \emph{pos}/\emph{neg} division boundary. AutoAssign~\cite{autoassign} tackles \emph{label assignment} in a fully data-driven manner by automatically determine the positives/negatives in both spatial and scale dimensions. 

These methods explore the optimal assigning strategy for individual objects, while failing to consider context information from a global perspective. DeTR~\cite{detr} examines the idea of global optimal matching. But the Hungarian algorithm they adopted can only work in a one-to-one assignment manner. So far, for the CNN based detectors in one-to-many scenarios, a global optimal assigning strategy remains uncharted.

\section{Method}

In this section, we first revisit the definition of the Optimal Transport problem and then demonstrate how we formulate the \emph{label assignment} in object detection into an OT problem. We also introduce two advanced designs which we suggest adopting to make the best use of OTA.

\subsection{Optimal Transport}\label{s3.1}

The Optimal Transport (OT) describes the following problem: supposing there are $m$ suppliers and $n$ demanders in a certain area. The $i$-th supplier holds $s_i$ units of goods while the $j$-th demander needs $d_j$ units of goods. Transporting cost for each unit of good from supplier $i$ to demander $j$ is denoted by $c_{ij}$. The goal of OT problem is to find a transportation plan $\pi^{*}=\{\pi_{i,j} | i=1,2,...m, j=1,2,...n\}$, according to which all goods from suppliers can be transported to demanders at a minimal transportation cost:

\begin{alignat}{2}
\begin{split}
\min_{\pi}\quad &\sum\limits_{i=1}^{m}\sum\limits_{j=1}^{n} c_{ij}\pi_{ij}. \\
\mbox{s.t.}\quad&\sum\limits_{i=1}^{m}\pi_{ij}=d_j,\quad \sum\limits_{j=1}^{n}\pi_{ij}=s_i, \\
&\sum\limits_{i=1}^{m}s_i=\sum\limits_{j=1}^{n}d_j, \\
&\pi_{ij}\geq 0,\quad i=1,2,...m, j=1,2,...n.
\end{split}\label{origin_formulation}
\end{alignat}

This is a linear program which can be solved in polynomial time. In our case, however, the resulting linear program is large, involving the square of feature dimensions with anchors in all scales. We thus address this issue by a fast iterative solution, named Sinkhorn-Knopp \cite{sinkhorn} (described in Appendix. \ref{sinkhorn}.)

\subsection{OT for Label Assignment}

In the context of object detection, supposing there are $m$ \emph{gt} targets and $n$ anchors (across all FPN~\cite{fpn} levels) for an input image $I$, we view each \emph{gt} as a supplier who holds $k$ units of positive labels (\emph{i.e.}, $s_i=k,i=1,2,...,m$), and each anchor as a demander who needs one unit of label (\emph{i.e.}, $d_j=1,j=1,2,...,n$). The cost $c^{fg}$ for transporting one unit of positive label from $gt_i$ to anchor $a_j$ is defined as the weighted summation of their $cls$ and $reg$ losses:
\begin{equation}
\begin{split}
c_{ij}^{fg} = & L_{cls}(P_{j}^{cls}(\theta), G_i^{cls}) + \\ & \alpha L_{reg}(P_{j}^{box}(\theta), G_i^{box}),
\end{split} \label{eq7}
\end{equation}
where $\theta$ stands for model‘s parameters. $P_{j}^{cls}$ and $P_{j}^{box}$ denote predicted $cls$ score and bounding box for $a_j$. $G_i^{cls}$ and $G_i^{box}$ denote ground truth class and bounding box for $gt$ $i$. $L_{cls}$ and $L_{reg}$ stand for cross entropy loss and IoU Loss~\cite{unitbox}. One can also replace these two losses with Focal Loss~\cite{focalloss} and GIoU~\cite{giou}/SmoothL1 Loss~\cite{fastrcnn}. $\alpha$ is the balanced coefficient.

Besides positive assigning, a large set of anchors are treated as negative samples during training. As the optimal transportation involves all anchors, we introduce another supplier -- \emph{background}, who only provides negative labels. In a standard OT problem, the total supply must be equal to the total demand. We thus set the number of negative labels that \emph{background} can supply as $n-m\times k$. The cost for transporting one unit of negative label from \emph{background} to $a_j$ is defined as:
\begin{equation}
\begin{split}
c_j^{bg} = L_{cls}(P_{j}^{cls}(\theta), \varnothing),
\end{split}
\end{equation}
where $\varnothing$ means the $background$ class. Concatenating this $c^{bg}\in \mathbb{R}^{1\times n}$ to the last row of $c^{fg}\in \mathbb{R}^{m\times n}$, we can get the complete form of the cost matrix $c\in \mathbb{R}^{(m+1)\times n}$. The supplying vector $s$ should be correspondingly updated as:
\begin{equation}
s_i=
\begin{cases}
k, & if\quad i\leq m \\
n-m\times k, & if\quad i=m+1 .
\end{cases}
\end{equation}

As we already have the cost matrix $c$, supplying vector $s\in \mathbb{R}^{m+1}$ and demanding vector $d\in \mathbb{R}^n$, the optimal transportation plan $\pi^* \in \mathbb{R}^{(m+1)\times n}$ can be obtained by solving this OT problem via the off-the-shelf Sinkhorn-Knopp Iteration \cite{sinkhorn}. After getting $\pi^*$, one can decode the corresponding label assigning solution by assigning each anchor to the supplier who transports the largest amount of labels to them. The subsequent processes (\emph{e.g.,} calculating losses based on assigning result, back-propagation) are exactly the same as in FCOS~\cite{fcos} and ATSS~\cite{atss}. Noted that the optimization process of OT problem only contains some matrix multiplications which can be accelerated by GPU devices, hence OTA \textbf{only increases the total training time by less than 20\%} and is totally cost-free in testing phase.

\begin{algorithm}[t]
\caption{Optimal Transport Assignment (OTA)}
\label{alg:Algorithm1}\label{ota}
\hspace*{0.02in} {\bf Input:}\\
\hspace*{0.2in} $I$ is an input image \\
\hspace*{0.2in} $A$ is a set of anchors \\
\hspace*{0.2in} $G$ is the $gt$ annotations for objects in image $I$ \\
\hspace*{0.2in} $\gamma$ is the regularization intensity in Sinkhorn-Knopp Iter. \\
\hspace*{0.2in} $T$ is the number of iterations in Sinkhorn-Knopp Iter. \\
\hspace*{0.2in} $\alpha$ is the balanced coefficient in Eq.~\ref{eq7} \\
\hspace*{0.02in} {\bf Output:} \\
\hspace*{0.2in} $\pi^*$ is the optimal assigning plan
\begin{algorithmic}[1]

\State $m$ $\leftarrow$ $\left\vert G \right\vert$, $n$ $\leftarrow$ $\left\vert A \right\vert$ \\
$P^{\text{cls}}, P^{\text{box}}$ $\leftarrow$ Forward($I$,$A$) \\
$s_i (i=1,2,...,m) \leftarrow$ Dynamic $k$ Estimation \\
$s_{m+1}\leftarrow n-\sum_{i=1}^{m}s_i$ \\
$d_j (j=1,2,...,n) \leftarrow$ OnesInit \\
pairwise $cls$ cost: $c_{\text{cls}}^{ij}$ = FocalLoss($P_j^{\text{cls}},G_i^{\text{cls}}$) \\
pairwise $reg$ cost: $c_{\text{reg}}^{ij}$ = IoULoss$(P_j^{\text{box}}$,$G_i^{\text{box}})$ \\
pairwise Center Prior cost: $c_{ij}^{\text{cp}}$ $\leftarrow$ $(A_j$, $G_i^{\text{box}})$ \\
$bg$ $cls$ cost: $c_{\text{cls}}^{\text{bg}}$ = FocalLoss($P_{j}^{\text{cls}}, \varnothing)$ \\
$fg$ cost: $c^{\text{fg}}$ = $c_{\text{cls}}+\alpha c_{\text{reg}}+c_{\text{cp}}$ \\
compute final cost matrix $c$ via concatenating $c_{\text{cls}}^{\text{bg}}$ to the last row of $c^{\text{fg}}$ \\
$v^0, u^0\leftarrow$ OnesInit
\For{t=0 \textbf{to} T}: \\
\hspace*{0.2in} $u^{t+1},v^{t+1}\leftarrow$ SinkhornIter($c,u^t,v^t,s,d$) 
\EndFor \\
compute optimal assigning plan $\pi^*$ according to Eq.~\ref{optimal_assigning_plan}
\State \Return $\pi^*$
\end{algorithmic}
\end{algorithm}

\subsection{Advanced Designs}\label{s3.2}

\paragraph{Center Prior.} Previous works~\cite{atss,paa,freeanchor} only select positive anchors from the center region of objects with limited areas, called \emph{Center Prior}. This is because they suffer from either a large number of ambiguous anchors or poor statistics in the subsequent process. Instead of relying on statistical characteristics, our OTA is based on global optimization methodology and thus is naturally resistant to these two issues. Theoretically, OTA can assign any anchor within the region of $gt$s' boxes as a positive sample.
However, for general detection datasets like COCO, we find the \emph{Center Prior} still benefit the training of OTA.
Forcing detectors focus on potential positive areas (~\emph{i.e.,} center areas) can help stabilize the training process, especially in the early stage of training, which will lead to a better final performance. Hence, we impose a \emph{Center Prior} to the cost matrix. For each $gt$, we select $r^2$ closest anchors from each FPN level according to the center distance between anchors and \emph{gt}s \footnote{For anchor-based methods, the distances are measured between the geometric center of anchors and $gt$s}. As for anchors not in the $r^2$ closest list, their corresponding entries in the cost matrix $c$ will be subject to an additional constant cost to reduce the possibility they are assigned as positive samples during the training stage. In Sec. \ref{exp}, we will demonstrate that although OTA adopts a certain degree of \emph{Center Prior} like other works~\cite{fcos,atss,freeanchor} do, OTA consistently outperforms counterparts by a large margin when $r$ is set to a large value (\emph{i.e.}, large number of potential positive anchors as well as more ambiguous anchors).

\paragraph{Dynamic $k$ Estimation.} Intuitively, the appropriate number of positive anchors for each $gt$ (\emph{i.e.}, $s_i$ in Sec.~\ref{s3.1}) should be different and based on many factors like objects' sizes, scales, and occlusion conditions, etc. As it is hard to directly model a mapping function from these factors to the positive anchor's number, we propose a simple but effective method to roughly estimate the appropriate number of positive anchors for each $gt$ based on the IoU values between predicted bounding boxes and $gt$s. Specifically, for each $gt$, we select the top $q$ predictions according to IoU values. These IoU values are summed up to represent this $gt$'s estimated number of positive anchors. We name this method as Dynamic $k$ Estimation. Such an estimation method is based on the following intuition: The appropriate number of positive anchors for a certain \emph{gt} should be positively correlated with the number of anchors that well-regress this \emph{gt}. In Sec. \ref{exp}, we present a detailed comparison between the fixed $k$ and Dynamic $k$ Estimation strategies.

A toy visualization of OTA is shown in Fig.~\ref{model}. We also describe the OTA's completed procedure including \emph{Center Prior} and Dynamic $k$ Estimation in Algorithm~\ref{ota}.

\section{Experiments} \label{exp}

In this section, we conduct extensive experiments on MS COCO 2017~\cite{coco} which contains about $118k$, $5k$ and $20k$ images for \emph{train}, \emph{val}, and \emph{test-dev} sets, respectively. For ablation studies, we train detectors on \emph{train} set and report the performance on \emph{val} set. Comparisons with other methods are conducted on \emph{test-dev} set. We also compare OTA with other methods on CrowdHuman~\cite{crowdhuman} validation set to demonstrate the superiority of OTA in crowd scenarios.

\subsection{Implementation Details}

If not specified, we use ResNet-50~\cite{resnet} pre-trained on ImageNet~\cite{imagenet} with FPN~\cite{fpn} as our default backbone. Most of experiments are trained with $90k$ iterations which is denoted as ``1$\times$''. The initial learning rate is 0.01 and is decayed by a factor of 10 after $60k$ and $80k$ iterations. Mini-batch size is set to 16. Following the common practice, the model is trained with SGD~\cite{sgd} on 8 GPUs.

OTA can be adopted in both anchor-based and anchor-free detectors, the following experiments are mainly conducted on FCOS~\cite{fcos} because of its simplicity. We adopt Focal Loss and IoU Loss as $L_{cls}$ and $L_{reg}$ that make up the cost matrix. $\alpha$ in Eq.~\ref{eq7} is set to $1.5$. For back-propagation, the regression loss is replaced by GIoU Loss and is re-weighted by a factor of $2$. IoU Branch is first introduced in YOLOv1~\cite{yolov1} and proved effective in modern one-stage object detectors by PAA~\cite{paa}. We also adopt IoU Branch as a default component in our experiments. The top $q$ in Sec.~\ref{s3.2} is directly set to 20, as we find this set of parameter values can consistently yield stable results in various situations.

\subsection{Ablation Studies and Analysis}

\paragraph{Effects of Individual Components.} We verify the effectiveness of each component in our proposed methods. For fair comparisons, all detectors' regression losses are multiplied by $2$, which is known as a useful trick to boost the AP at high IoU thresholds~\cite{libra}. As seen in Table~\ref{component}, when no auxiliary branch is adopted, OTA outperforms FCOS by 0.9\% AP (39.2\%~\emph{v.s.}38.3\%). This gap almost remains the same after adding IoU branch to both of them (39.5\%~\emph{v.s.} 40.3\% and 38.8\%~\emph{v.s.} 39.6\% with or without center prior, respectively). Finally, dynamic $k$ pushes AP to a new state-of-the-art 40.7\%. In the whole paper, we emphasize that \textbf{OTA can be applied to both anchor-based and anchor-free detectors}. Hence we also adopt OTA on RetinaNet~\cite{focalloss} with only one square anchor per-location across feature maps. As shown in Table~\ref{component}, the AP values of OTA-FCOS and OTA-RetinaNet are exactly the same, demonstrating OTA's applicability on both anchor-based and anchor-free detectors.

\begin{table}[]
\setlength{\tabcolsep}{0.6mm}{
\begin{tabular}{c|c|cc|ccc}
\hline
Method                & Aux. Branch & Center        & Dyn. $k$  & AP & AP$_{50}$ & AP$_{75}$ \\ \hline
\multirow{4}{*}{FCOS} &     -       & \checkmark &            & 38.3 & 57.1 & 41.3    \\
                      & CenterNess  & \checkmark &            & 38.9 & 57.5 & 42.0  \\
                    &  IoU        &            &            &  38.8 & 57.7     & 41.8    \\
                      & IoU         & \checkmark &            & 39.5 & 57.6 & 42.9  \\ \hline
\multirow{4}{*}{\begin{tabular}[c]{@{}c@{}}OTA\\(FCOS)\end{tabular}}  & -           & \checkmark &            & 39.2 & 58.3 & 42.2  \\ 
                      & IoU         &            &            & 39.6 & 58.1 & 42.5    \\
                      & IoU         & \checkmark &            & 40.3  &   \textbf{58.6}   &  43.7    \\
                      & IoU         & \checkmark & \checkmark & \textbf{40.7}    &  58.4    &   \textbf{44.3}   \\ \hline
\begin{tabular}[c]{@{}c@{}}OTA\\(RetinaNet) \end{tabular}  &  IoU  &  \checkmark    &  \checkmark    &   \textbf{40.7}  &\textbf{58.6} &   44.1  \\ \hline
\end{tabular}}
    \caption{Ablation studies on each components in OTA. ``Center'' stands for \emph{Center Prior}  and \emph{Center Sampling} for OTA and FCOS, respectively. Dyn.$k$ is the abbreviation of our proposed Dynamic $k$ Estimation strategy.}\label{component}
\end{table}


\paragraph{Effects of $r$.} The values of radius $r$ for \emph{Center Prior} serve to control the number of candidate anchors for each $gt$. If adopting a small $r$, only anchors near objects' centers could be assigned as positives, helping the optimization process focus on regions that are more likely to be informative. As $r$ increases, the number of candidates also quadratically increases, leading to potential instability in the optimization process. For example, when $r$ is set to $3$, $5$ or $7$, their corresponding numbers of candidate anchors are $45$, $125$ and $245$\footnote{Total number of potential positive anchors equals to ($r^2*$FPN Levels).}, respectively. We study behaviors of ATSS~\cite{atss}, PAA~\cite{paa}, and OTA under different values of $r$ in Table~\ref{r_ablation}. OTA achieves the best performance (40.7\% AP) when $r$ is set to $5$. When $r$ is set to 3 as ATSS and PAA do, OTA also achieves 40.6\% AP, indicating that most potential positive anchors are near the center of objects on COCO. While $r$ is set to 7, the performance only slightly drops 0.3\%, showing that OTA is insensitive to the hyper-parameter $r$.

\begin{figure}[!t]
\includegraphics[width=8.3cm]{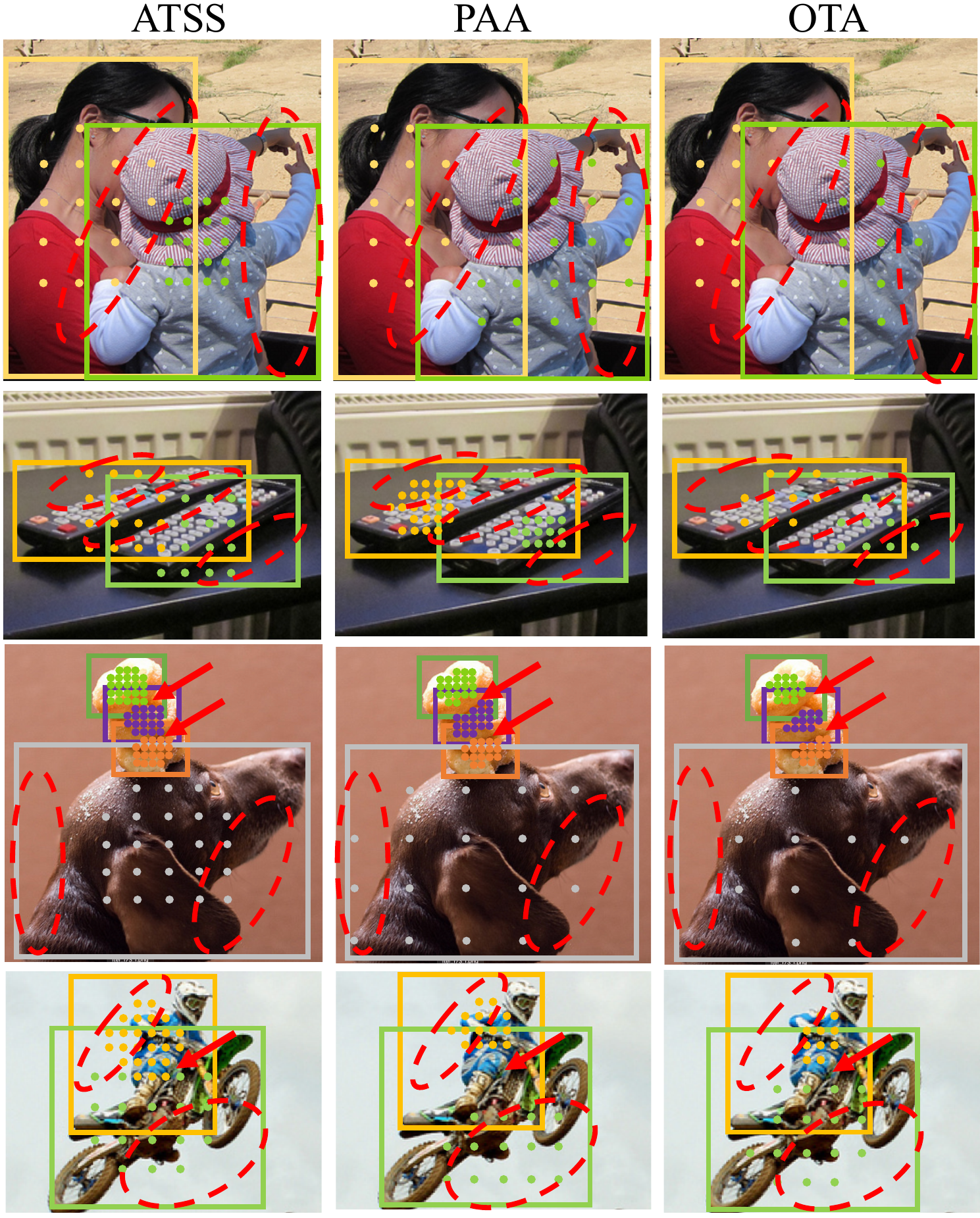}
   \caption{Visualizations of assigning results. For PAA, the dots stand for geometric centers of positive anchor boxes. For ATSS and OTA, the dots stand for positive anchor points. Rectangles represent the $gt$ bounding boxes. To clearly illustrate the differences between different assigning strategies, we set $r$ to 5 for all methods. Only FPN layers with the largest number of positive anchors are shown for better visualization.}
\label{pos_vis}
\end{figure}

\paragraph{Ambiguous Anchors Handling.} Most existing dynamic label assigning methods~\cite{atss,paa,freeanchor} only conduct a small candidate set for each $gt$, because a large number of candidates brings trouble -- when occlusion happens or several objects are close enough, an anchor may simultaneously be a qualified candidate for multiple $gt$s. We define such anchors as \emph{ambiguous anchors}. Previous methods mainly handle this ambiguity by introducing hand-crafted rules ~\emph{e.g., Min Area~\cite{fcos}, Max IoU~\cite{atss,paa,focalloss} and Min Loss\footnote{Assigning ambiguous anchor to the $gt$ with the minimal loss.}}. To illustrate OTA's superiority on ambiguous handling, We count the number of ambiguous anchors in ATSS, PAA and OTA, and evaluate their corresponding performance under different $r$ in Table~\ref{r_ablation}. Noted that the optimal assigning plan in OTA is continuous, hence we define anchor $a_j$ as an ambiguous anchor if $\max {\pi^*_j}<0.9$. Table~\ref{r_ablation} shows that for ATSS, the number of ambiguous anchors greatly increases as $r$ varies from 3 to 7. Its performance correspondingly drops from 39.4\% to 37.2\%. For PAA, the number of ambiguous anchors is less sensitive to $r$, but its performance still drops 0.8\%, indicating that \emph{Max IoU} adopted by PAA is not an ideal prior to ambiguous anchors. In OTA, when multiple $gt$s tend to transport positive labels to the same anchor, the OT algorithm will automatically resolve their conflicts based on the principle of minimum global costs. Hence the number of ambiguous anchor for OTA remains low and barely increases as $r$ increases from 3 to 7. The corresponding performance is also stable. 

Further, we \emph{manually} assign the ambiguous anchors based on hand-crafted rules before performing OTA. In this case, OTA is only in charge of \emph{pos}/\emph{neg} samples division. Table~\ref{bb} shows that such a combination of hand-crafted rules and OTA decreases the AP by 0.7\% and 0.4\%, respectively. Finally, we visualize some assigning results in Fig.~\ref{pos_vis}. Red arrows and dashed ovals highlight the ambiguous regions (\emph{i.e.,} overlaps between different $fg$s or junctions between $fg$s and $bg$). Suffering from the lack of context and global information, ATSS and PAA perform poorly in such regions, leading to sub-optimal detection performances. Conversely, OTA assigns much less positive anchors in such regions, which we believe is a desired behavior.

\begin{table}[!t]
\centering
\setlength{\tabcolsep}{0.8mm}{
\begin{tabular}{c|ccc|ccc|ccc}
\hline
Method & \multicolumn{3}{c|}{ATSS~\cite{atss}} & \multicolumn{3}{c|}{PAA~\cite{paa}} & \multicolumn{3}{c}{OTA} \\ \hline
$r$  & 3       & 5      & 7      & 3      & 5      & 7      & 3      & 5      & 7     \\ \hline
$N_{amb.}$  &    2.1    &  15.9  &  36.3  &  0.5  &  0.8  &  1.2  & 0.2  &  0.2 & 0.3 \\ \hline
AP   &  39.4       &   38.0     &    37.2   &    40.3      &  40.1     &   39.5    &  40.6   &   \textbf{40.7}   &    40.4    \\ 
AP$_{50}$ &  57.5       &   56.7     &   55.8     &   \textbf{58.9}     &   58.4     &   57.5        &    58.7    &    58.4    &  58.3     \\ 
AP$_{75}$ &  42.7       &   40.4    &   39.8     &    43.4    &   43.4      &   42.4     &    44.1    &    \textbf{44.3}    &  43.6     \\ \hline
\end{tabular}}
    \caption{Performances of different label assigning strategies under different number of anchor candidates. $N_{amb.}$ denotes the average number of ambiguous anchors per-image calculated on COCO \emph{train} set.}\label{r_ablation}
\end{table}

\begin{table}[!h]
\centering
\setlength{\tabcolsep}{1.0mm}{
\begin{tabular}{l|ccc}
\hline
Method & AP & AP$_{50}$ & AP$_{75}$  \\ \hline
Min Area~\cite{fcos} \emph{f.b.} OTA              & 40.0   & 57.8  & 43.6         \\
Max IoU~\cite{atss} \emph{f.b.} OTA               & 40.3   & 58.1  & 43.7             \\
Min Loss \emph{f.b.} OTA & 40.3   & 57.9  & 43.6 \\ \hline
OTA       & \textbf{40.7}   & \textbf{58.4}  & \textbf{44.3}          \\ \hline
\end{tabular}}
\caption{Performance comparisons on ambiguity handling between OTA and other human-designed strategies on the COCO \emph{val} set.. \emph{f.b.} denotes ``followed by''.}\label{bb}
\end{table}

\paragraph{Effects of $k$.} Before performing Sinkhorn-Knopp Iteration, we need to define how many positive labels can each $gt$ supply. This value also represents how many anchors every $gt$ needs for better convergence. A naive way is setting $k$ to a constant value for all $gt$s. We try different values of $k$ from 1 to 20. As seen in Table~\ref{dynk}, among all different values, $k$=10 and $k$=12 achieve the best performances. As $k$ increases from 10 to 20, the possibility that an anchor is suitable as a positive sample for two close targets at the same time also increases, but there is no obvious performance drop (0.2\%) according to Table~\ref{dynk} which proves OTA's superiority in handling potential ambiguity. When $k$=1, OTA becomes a one-to-one assigning strategy, the same as in DeTR. The poor performance tells us that achieving competitive performance via one-to-one assignment under the 1$\times$ scheduler remains challenging, unless an auxiliary one-to-many supervision is added~\cite{wang2020end}.

\begin{table}[!h]
\centering
\begin{tabular}{c|c|ccccc}
\hline
$k$  & AP & AP$_{50}$ & AP$_{75}$ & APs & APm & APl \\ \hline
1    & 36.5   &55.4  &38.8      & 21.4    &39.7 & 46.2  \\
5    & 39.5   &58.1  &42.7      & 23.1    &43.0 & 50.6  \\
8    & 39.8  & 58.4     & 42.9     & 22.7    & 43.6    & 51.5  \\
10   & 40.3   &  \textbf{58.6}   & 43.7 & 23.4  & 44.2 & 52.1   \\
12   & 40.3   &  \textbf{58.6}   &  43.6    &  23.2   &  44.2   &  51.9  \\
15   & 40.2   &  58.4   &  43.6    &  23.2   &  44.1   &  51.9  \\
20   & 40.1   &  58.2   &  43.6    &  \textbf{23.5}     &44.0    &  52.8 \\ \hline
Dyn. $k$ & \textbf{40.7}& 58.4 &\textbf{44.3}  &23.2 &\textbf{45.0} &\textbf{53.6}   \\ \hline
\end{tabular}
\caption{Analysis of different values of $k$ and Dynamic $k$ Estimation strategy on the COCO \emph{val} set.}\label{dynk}
\end{table}

Fixing $k$ strategy assumes every $gt$ has the same number of appropriate positive anchors. However, we believe that this number for each $gt$ should vary and may be affected by many factors like objects' sizes, spatial attitudes, and occlusion conditions, etc. Hence we adopt the Dynamic $k$ Estimation proposed in Sec~\ref{s3.2} and compare its performance to the fixed $k$ strategy. Results in Table~\ref{dynk} shows that dynamic $k$ surpasses the best performance of fixed $k$ by 0.4\% AP, validating our point and the effectiveness of Dynamic $k$ Estimation strategy.

\subsection{Comparison with State-of-the-art Methods.}

We compare our final models with other state-of-the-art one-stage detectors on MS COCO \emph{test-dev}. Following previous works~\cite{focalloss,fcos}, we randomly scale the shorter side of images in the range from 640 to 800. Besides, we double the total number of iterations to $180K$ with the learning rate change points scaled proportionally. Other settings are consistent with~\cite{focalloss,fcos}.

As shown in Table~\ref{sota}, our method with ResNet-101-FPN achieves 45.3\% AP, outperforms all other methods with the same backbone including ATSS (43.6\% AP), AutoAssign (44.5\% AP) and PAA (44.6\% AP). Noted that for PAA, we remove the \emph{score voting} procedure for fair comparisons between different label assigning strategies. With ResNeXt-64x4d-101-FPN~\cite{resnext}, the performance of OTA can be further improved to 47.0\% AP. To demonstrate the compatibility of our method with other advanced technologies in object detection, we adopt Deformable Convolutional Networks (DCN)~\cite{dcnv2} to ResNeXt backbones as well as the last convolution layer in the detection head. This improves our model's performance from 47.0\% AP to 49.2\% AP. Finally, with the multi-scale testing technique, our best model achieves 51.5\% AP.

\begin{table*}[!t]
\centering
\scalebox{0.9}{
\resizebox{\textwidth}{!}{
\begin{tabular}{l|c|c|ccc|ccc}
\hline
Method      & Iteration & Backbone          & AP   & AP$_{50}$ & AP$_{75}$ & APs  & APm  & APl  \\ \hline
RetinaNet~\cite{focalloss}   & 135k      & ResNet-101         & 39.1 & 59.1      & 42.3      & 21.8 & 42.7 & 50.2 \\
FCOS~\cite{fcos}        & 180k      & ResNet-101         & 41.5 & 60.7      & 45.0      & 24.4 & 44.8 & 51.6 \\
NoisyAnchor~\cite{noisyanchors} & 180k      & ResNet-101         & 41.8 & 61.1      & 44.9      & 23.4 & 44.9 & 52.9 \\
FreeAnchor~\cite{freeanchor}  & 180k      & ResNet-101         & 43.1 & 62.2      & 46.4      & 24.5 & 46.1 & 54.8 \\
SAPD~\cite{sapd}        & 180k      & ResNet-101         & 43.5 & 63.6      & 46.5      & 24.9 & 46.8 & 54.6 \\
MAL~\cite{metaanchor}         & 180k      & ResNet-101         & 43.6 & 61.8      & 47.1      & 25.0 & 46.9 & 55.8 \\
ATSS~\cite{atss}        & 180k      & ResNet-101         & 43.6 & 62.1      & 47.4      & 26.1 & 47.0 & 53.6 \\
AutoAssign~\cite{autoassign}  & 180k      & ResNet-101         & 44.5 & \textbf{64.3}      & 48.4      & 25.9 & 47.4 & 55.0 \\
PAA~\cite{paa}  & 180k  & ResNet-101 & 44.6 & 63.3   & 48.4  & 26.4 & 48.5 & 56.0 \\
OTA (Ours)        & 180k      & ResNet-101         & \textbf{45.3} & 63.5  & \textbf{49.3}      & \textbf{26.9} & \textbf{48.8} & \textbf{56.1} \\ \hline
FoveaBox~\cite{foveabox}    & 180k      & ResNeXt-101       & 42.1 & 61.9      & 45.2      & 24.9 & 46.8 & 55.6 \\
FSAF~\cite{fsaf}        & 180k      & ResNeXt-64x4d-101 & 42.9 & 63.8      & 46.3      & 26.6 & 46.2 & 52.7 \\
FCOS~\cite{fcos}        & 180k      & ResNeXt-64x4d-101 & 43.2 & 62.8      & 46.6      & 26.5 & 46.2 & 53.3 \\
NoisyAnchor~\cite{noisyanchors} & 180k      & ResNeXt-101       & 44.1 & 63.8      & 47.5      & 26.0 & 47.4 & 55.0 \\
FreeAnchor~\cite{freeanchor}  & 180k      & ResNeXt-64x4d-101 & 44.9 & 64.3      & 48.5      & 26.8 & 48.3 & 55.9 \\
SAPD~\cite{sapd}        & 180k      & ResNeXt-64x4d-101 & 45.4 & 65.6      & 48.9      & 27.3 & 48.7 & 56.8 \\
ATSS~\cite{atss}        & 180k      & ResNeXt-64x4d-101 & 45.6 & 64.6      & 49.7      & 28.5 & 48.9 & 55.6 \\
MAL~\cite{metaanchor}         & 180k      & ResNeXt101        & 45.9 & 65.4      & 49.7      & 27.8 & 49.1 & 57.8 \\
AutoAssign~\cite{autoassign}  & 180k      & ResNeXt-64x4d-101 & 46.5 & \textbf{66.5}      & 50.7      & 28.3 & 49.7 & 56.6 \\
PAA~\cite{paa}         & 180k      & ResNeXt-64x4d-101 & 46.6 & 65.6 & 50.7 &28.7  & \textbf{50.5} & \textbf{58.1} \\
OTA (Ours)        & 180k      & ResNeXt-64x4d-101 & \textbf{47.0} & 65.8 & \textbf{51.1} & \textbf{29.2} & 50.4 & 57.9      \\ \hline
SAPD~\cite{sapd}        & 180k      & ResNeXt-64x4d-101-DCN & 47.4 & 67.4      & 51.1      & 28.1 & 50.3 & 61.5 \\
ATSS~\cite{atss}      & 180k      & ResNeXt-64x4d-101-DCN & 47.7 & 66.5      & 51.9      & 29.7 & 50.8 & 59.4 \\
AutoAssign~\cite{autoassign}  & 180k      & ResNeXt-64x4d-101-DCN & 48.3 & 67.4      & 52.7      & 29.2 & 51.0 & 60.3 \\
PAA~\cite{paa}      & 180k      & ResNeXt-64x4d-101-DCN & 48.6 & 67.5 & 52.7 & 29.9 & 52.2 & 61.5 \\
OTA (Ours)   & 180k      & ResNeXt-64x4d-101-DCN & \textbf{49.2} & \textbf{67.6} & \textbf{53.5} & \textbf{30.0} & \textbf{52.5} & \textbf{62.3} \\ \hline
ATSS~\cite{atss}$^*$        & 180k      & ResNeXt-64x4d-101-DCN & 50.7 & \textbf{68.9} & 56.3 & 33.2 & 52.9 & 62.2 \\
PAA~\cite{paa}$^*$         & 180k      & ResNeXt-64x4d-101-DCN & 51.3 & 68.8 & 56.6 & \textbf{34.3} & 53.5 & 63.6 \\
OTA (Ours)$^*$    & 180k      & ResNeXt-64x4d-101-DCN & \textbf{51.5} & 68.6 & \textbf{57.1} & 34.1  & \textbf{53.7}  & \textbf{64.1}  \\ \hline
\end{tabular}}}
\caption{Performance comparison with state-of-the-art one-stage detectors on MS COCO 2017 \emph{test-dev} set. * indicates the specific form of multi-scale testing that adopted in ATSS~\cite{atss}.}\label{sota}
\end{table*}

\subsection{Experiments on CrowdHuman}

\begin{table}[]
\centering
\begin{tabular}{l|ccc}
\hline
Method     & MR   & AP   & Recall \\ \hline
Faster R-CNN \emph{with} FPN~\cite{fpn} &48.7 &86.1 & 90.4   \\ \hline
RetinaNet~\cite{focalloss} & 58.8 & 81.0 & 88.2   \\
FCOS~\cite{fcos}       & 55.0 & 86.4 & 94.1   \\ \hline
FreeAnchor~\cite{freeanchor} & 51.3 & 83.9 & 89.8   \\
ATSS~\cite{atss}       & 49.5 & 87.4 & 94.2   \\
PAA~\cite{paa}        & 52.2 & 86.0 & 92.0   \\
LLA~\cite{lla}        & 47.9 & 88.0 & 94.0   \\\hline
OTA~(Ours)        & \textbf{46.6} & \textbf{88.4} & \textbf{95.1}   \\ \hline
\end{tabular}
\caption{Performance comparison on the CrowdHuman validation set. All experiments are conducted under 2.5x scheduler.}\label{crowdhuman}
\end{table}

Object detection in crowded scenarios has raised more and more attention~\cite{liu2019adaptive, r2nms, psrcnn, lla}. Compared to dataset designed for general object detection like COCO, ambiguity happens more frequently in crowded dataset. Hence to demonstrate OTA's advantage on handling ambiguous anchors, it is necessary to conduct experiments on a crowded dataset -- Crowdhuman~\cite{crowdhuman}. CrowdHuman contains 15000, 4370, and 5000 images in training, validation, and test set, respectively, with the average number of persons in an image 22.6. For all experiments, we train the detectors for 30 epochs (\emph{i.e.}, 2.5x) for better convergence. NMS threshold is set to 0.6. We adopt ResNet-50~\cite{resnet} as the default backbone in our experiments. Other settings are the same as our experiments on COCO. For evaluation, we follow the standard Caltech~\cite{caltech} evaluation metric -- MR, which stands for the Log-Average Missing Rate over false positives per image (FPPI) ranging in $[10^{-2},10^0]$. AP and Recall are also reported for reference. All evaluation results are reported on the CrowdHuman \emph{val} subset.

As shown in Table~\ref{crowdhuman}, RetinaNet and FCOS only achieve 58.8\% and 55.0\% MR respectively, which are far worse than two stage detectors like Faster R-CNN (with FPN), revealing the dilemma of one-stage detectors in crowd scenarios. Starting from FreeAnchor, the performances of one-stage detectors gradually get improved by the dynamic label assigning strategies. ATSS achieves 49.5\% MR, which is very close to the performance of Faster R-CNN (48.7\% AP). Recent proposed LLA~\cite{lla} leverages loss-aware label assignment, which is similar to OTA and achieves 47.9\% MR. However, our OTA takes a step forward by introducing global information into the label assignment, boosting MR to 46.6\%. The AP and Recall of OTA also surpass other existing one-stage detectors by a clear margin. 

Although PAA achieves competitive performance with OTA on COCO, it performs struggling on CrowdHuman. We conjecture that PAA needs clear \emph{pos}/\emph{neg} decision boundaries to help GMM learn better clusters. But in crowded scenarios, such clear boundaries may not exist because potential negative samples usually cover a sufficient amount of foreground areas, resulting in PAA's poor performance. Also, PAA performs per-\emph{gt}'s clustering, which heavily increases the training time on crowded datasets like CrowdHuman. Compared to PAA, OTA still shows promising results, which demonstrates OTA's superiority on various detection benchmarks.

\section{Conclusion}

In this paper, we propose Optimal Transport Assignment (OTA) -- an optimization theory based label assigning strategy. OTA formulates the label assigning procedure in object detection into an Optimal Transport problem, which aims to transport labels from ground-truth objects and backgrounds to anchors at minimal transporting costs. To determine the number of positive labels needed by each $gt$, we further propose a simple estimation strategy based on the IoU values between predicted bounding boxes and each $gt$. As shown in experiments, OTA achieves the new SOTA performance on MS COCO. Because OTA can well-handle the assignment of ambiguous anchors, it also outperforms all other one-stage detectors on CrowdHuman dataset by a large margin, demonstrating its strong generalization ability.

\section*{Acknowledgements}

This research was partially supported by National Key R\&D Program of China (No. 2017YFA0700800), and Beijing Academy of Artiﬁcial Intelligence (BAAI).

{\small
\bibliographystyle{ieee_fullname}
\bibliography{egpaper_final}

\begin{thebibliography}{10}\itemsep=-1pt

\bibitem{sgd}
L{\'e}on Bottou.
\newblock Large-scale machine learning with stochastic gradient descent.
\newblock In {\em Proceedings of COMPSTAT'2010}, pages 177--186. Springer,
  2010.

\bibitem{cascadercnn}
Zhaowei Cai and Nuno Vasconcelos.
\newblock Cascade r-cnn: Delving into high quality object detection.
\newblock In {\em Proceedings of the IEEE conference on computer vision and
  pattern recognition}, pages 6154--6162, 2018.

\bibitem{detr}
Nicolas Carion, Francisco Massa, Gabriel Synnaeve, Nicolas Usunier, Alexander
  Kirillov, and Sergey Zagoruyko.
\newblock End-to-end object detection with transformers.
\newblock {\em arXiv preprint arXiv:2005.12872}, 2020.

\bibitem{reppointsv2}
Yihong Chen, Zheng Zhang, Yue Cao, Liwei Wang, Stephen Lin, and Han Hu.
\newblock Reppoints v2: Verification meets regression for object detection.
\newblock {\em arXiv preprint arXiv:2007.08508}, 2020.

\bibitem{sinkhorn}
Marco Cuturi.
\newblock Sinkhorn distances: Lightspeed computation of optimal transport.
\newblock In {\em Advances in neural information processing systems}, pages
  2292--2300, 2013.

\bibitem{imagenet}
Jia Deng, Wei Dong, Richard Socher, Li-Jia Li, Kai Li, and Li Fei-Fei.
\newblock Imagenet: A large-scale hierarchical image database.
\newblock In {\em 2009 IEEE conference on computer vision and pattern
  recognition}, pages 248--255. Ieee, 2009.

\bibitem{caltech}
Piotr Doll{\'a}r, Christian Wojek, Bernt Schiele, and Pietro Perona.
\newblock Pedestrian detection: A benchmark.
\newblock In {\em 2009 IEEE Conference on Computer Vision and Pattern
  Recognition}, pages 304--311. IEEE, 2009.

\bibitem{centernet}
Kaiwen Duan, Song Bai, Lingxi Xie, Honggang Qi, Qingming Huang, and Qi Tian.
\newblock Centernet: Keypoint triplets for object detection.
\newblock In {\em Proceedings of the IEEE International Conference on Computer
  Vision}, pages 6569--6578, 2019.

\bibitem{psrcnn}
Zheng Ge, Zequn Jie, Xin Huang, Rong Xu, and Osamu Yoshie.
\newblock Ps-rcnn: Detecting secondary human instances in a crowd via primary
  object suppression.
\newblock In {\em 2020 IEEE International Conference on Multimedia and Expo
  (ICME)}, pages 1--6. IEEE, 2020.

\bibitem{lla}
Zheng Ge, Jianfeng Wang, Xin Huang, Songtao Liu, and Osamu Yoshie.
\newblock Lla: Loss-aware label assignment for dense pedestrian detection.
\newblock {\em arXiv preprint arXiv:2101.04307}, 2021.

\bibitem{fastrcnn}
Ross Girshick.
\newblock Fast r-cnn.
\newblock In {\em Proceedings of the IEEE international conference on computer
  vision}, pages 1440--1448, 2015.

\bibitem{maskrcnn}
Kaiming He, Georgia Gkioxari, Piotr Doll{\'a}r, and Ross Girshick.
\newblock Mask r-cnn.
\newblock In {\em Proceedings of the IEEE international conference on computer
  vision}, pages 2961--2969, 2017.

\bibitem{resnet}
Kaiming He, Xiangyu Zhang, Shaoqing Ren, and Jian Sun.
\newblock Deep residual learning for image recognition.
\newblock In {\em Proceedings of the IEEE conference on computer vision and
  pattern recognition}, pages 770--778, 2016.

\bibitem{densebox}
Lichao Huang, Yi Yang, Yafeng Deng, and Yinan Yu.
\newblock Densebox: Unifying landmark localization with end to end object
  detection.
\newblock {\em arXiv preprint arXiv:1509.04874}, 2015.

\bibitem{r2nms}
Xin Huang, Zheng Ge, Zequn Jie, and Osamu Yoshie.
\newblock Nms by representative region: Towards crowded pedestrian detection by
  proposal pairing.
\newblock In {\em Proceedings of the IEEE/CVF Conference on Computer Vision and
  Pattern Recognition}, pages 10750--10759, 2020.

\bibitem{paa}
Kang Kim and Hee~Seok Lee.
\newblock Probabilistic anchor assignment with iou prediction for object
  detection.
\newblock {\em arXiv preprint arXiv:2007.08103}, 2020.

\bibitem{foveabox}
Tao Kong, Fuchun Sun, Huaping Liu, Yuning Jiang, Lei Li, and Jianbo Shi.
\newblock Foveabox: Beyound anchor-based object detection.
\newblock {\em IEEE Transactions on Image Processing}, 29:7389--7398, 2020.

\bibitem{cornernet}
Hei Law and Jia Deng.
\newblock Cornernet: Detecting objects as paired keypoints.
\newblock In {\em Proceedings of the European Conference on Computer Vision
  (ECCV)}, pages 734--750, 2018.

\bibitem{noisyanchors}
Hengduo Li, Zuxuan Wu, Chen Zhu, Caiming Xiong, Richard Socher, and Larry~S
  Davis.
\newblock Learning from noisy anchors for one-stage object detection.
\newblock In {\em Proceedings of the IEEE/CVF Conference on Computer Vision and
  Pattern Recognition}, pages 10588--10597, 2020.

\bibitem{fpn}
Tsung-Yi Lin, Piotr Doll{\'a}r, Ross Girshick, Kaiming He, Bharath Hariharan,
  and Serge Belongie.
\newblock Feature pyramid networks for object detection.
\newblock In {\em Proceedings of the IEEE conference on computer vision and
  pattern recognition}, pages 2117--2125, 2017.

\bibitem{focalloss}
Tsung-Yi Lin, Priya Goyal, Ross Girshick, Kaiming He, and Piotr Doll{\'a}r.
\newblock Focal loss for dense object detection.
\newblock In {\em Proceedings of the IEEE international conference on computer
  vision}, pages 2980--2988, 2017.

\bibitem{coco}
Tsung-Yi Lin, Michael Maire, Serge Belongie, James Hays, Pietro Perona, Deva
  Ramanan, Piotr Doll{\'a}r, and C~Lawrence Zitnick.
\newblock Microsoft coco: Common objects in context.
\newblock In {\em European conference on computer vision}, pages 740--755.
  Springer, 2014.

\bibitem{rfb}
Songtao Liu, Di Huang, et~al.
\newblock Receptive field block net for accurate and fast object detection.
\newblock In {\em Proceedings of the European Conference on Computer Vision
  (ECCV)}, pages 385--400, 2018.

\bibitem{liu2019adaptive}
Songtao Liu, Di Huang, and Yunhong Wang.
\newblock Adaptive nms: Refining pedestrian detection in a crowd.
\newblock In {\em Proceedings of the IEEE/CVF Conference on Computer Vision and
  Pattern Recognition}, pages 6459--6468, 2019.

\bibitem{asff}
Songtao Liu, Di Huang, and Yunhong Wang.
\newblock Learning spatial fusion for single-shot object detection.
\newblock {\em arXiv preprint arXiv:1911.09516}, 2019.

\bibitem{pat}
Songtao Liu, Di Huang, and Yunhong Wang.
\newblock Pay attention to them: deep reinforcement learning-based cascade
  object detection.
\newblock {\em IEEE transactions on neural networks and learning systems},
  31(7):2544--2556, 2019.

\bibitem{ssd}
Wei Liu, Dragomir Anguelov, Dumitru Erhan, Christian Szegedy, Scott Reed,
  Cheng-Yang Fu, and Alexander~C Berg.
\newblock Ssd: Single shot multibox detector.
\newblock In {\em European conference on computer vision}, pages 21--37.
  Springer, 2016.

\bibitem{libra}
Jiangmiao Pang, Kai Chen, Jianping Shi, Huajun Feng, Wanli Ouyang, and Dahua
  Lin.
\newblock Libra r-cnn: Towards balanced learning for object detection.
\newblock In {\em Proceedings of the IEEE conference on computer vision and
  pattern recognition}, pages 821--830, 2019.

\bibitem{borderdet}
Han Qiu, Yuchen Ma, Zeming Li, Songtao Liu, and Jian Sun.
\newblock Borderdet: Border feature for dense object detection.
\newblock In {\em European Conference on Computer Vision}, pages 549--564.
  Springer, 2020.

\bibitem{yolov1}
Joseph Redmon, Santosh Divvala, Ross Girshick, and Ali Farhadi.
\newblock You only look once: Unified, real-time object detection.
\newblock In {\em Proceedings of the IEEE conference on computer vision and
  pattern recognition}, pages 779--788, 2016.

\bibitem{yolo9000}
Joseph Redmon and Ali Farhadi.
\newblock Yolo9000: better, faster, stronger.
\newblock In {\em Proceedings of the IEEE conference on computer vision and
  pattern recognition}, pages 7263--7271, 2017.

\bibitem{yolov3}
Joseph Redmon and Ali Farhadi.
\newblock Yolov3: An incremental improvement.
\newblock {\em arXiv preprint arXiv:1804.02767}, 2018.

\bibitem{fasterrcnn}
Shaoqing Ren, Kaiming He, Ross Girshick, and Jian Sun.
\newblock Faster r-cnn: Towards real-time object detection with region proposal
  networks.
\newblock In {\em Advances in neural information processing systems}, pages
  91--99, 2015.

\bibitem{giou}
Hamid Rezatofighi, Nathan Tsoi, JunYoung Gwak, Amir Sadeghian, Ian Reid, and
  Silvio Savarese.
\newblock Generalized intersection over union: A metric and a loss for bounding
  box regression.
\newblock In {\em Proceedings of the IEEE Conference on Computer Vision and
  Pattern Recognition}, pages 658--666, 2019.

\bibitem{crowdhuman}
Shuai Shao, Zijian Zhao, Boxun Li, Tete Xiao, Gang Yu, Xiangyu Zhang, and Jian
  Sun.
\newblock Crowdhuman: A benchmark for detecting human in a crowd.
\newblock {\em arXiv preprint arXiv:1805.00123}, 2018.

\bibitem{dynhead}
Lin Song, Yanwei Li, Zhengkai Jiang, Zeming Li, Hongbin Sun, Jian Sun, and
  Nanning Zheng.
\newblock Fine-grained dynamic head for object detection.
\newblock {\em arXiv preprint arXiv:2012.03519}, 2020.

\bibitem{treev2}
Lin Song, Yanwei Li, Zhengkai Jiang, Zeming Li, Xiangyu Zhang, Hongbin Sun,
  Jian Sun, and Nanning Zheng.
\newblock Rethinking learnable tree filter for generic feature transform.
\newblock {\em arXiv preprint arXiv:2012.03482}, 2020.

\bibitem{fcos}
Zhi Tian, Chunhua Shen, Hao Chen, and Tong He.
\newblock Fcos: Fully convolutional one-stage object detection.
\newblock In {\em Proceedings of the IEEE international conference on computer
  vision}, pages 9627--9636, 2019.

\bibitem{transformer}
Ashish Vaswani, Noam Shazeer, Niki Parmar, Jakob Uszkoreit, Llion Jones,
  Aidan~N Gomez, {\L}ukasz Kaiser, and Illia Polosukhin.
\newblock Attention is all you need.
\newblock In {\em Advances in neural information processing systems}, pages
  5998--6008, 2017.

\bibitem{guidedanchor}
Jiaqi Wang, Kai Chen, Shuo Yang, Chen~Change Loy, and Dahua Lin.
\newblock Region proposal by guided anchoring.
\newblock In {\em Proceedings of the IEEE Conference on Computer Vision and
  Pattern Recognition}, pages 2965--2974, 2019.

\bibitem{wang2020end}
Jianfeng Wang, Lin Song, Zeming Li, Hongbin Sun, Jian Sun, and Nanning Zheng.
\newblock End-to-end object detection with fully convolutional network.
\newblock {\em arXiv preprint arXiv:2012.03544}, 2020.

\bibitem{mpsr}
Jiaxi Wu, Songtao Liu, Di Huang, and Yunhong Wang.
\newblock Multi-scale positive sample refinement for few-shot object detection.
\newblock In {\em European Conference on Computer Vision}, pages 456--472.
  Springer, 2020.

\bibitem{resnext}
Saining Xie, Ross Girshick, Piotr Doll{\'a}r, Zhuowen Tu, and Kaiming He.
\newblock Aggregated residual transformations for deep neural networks.
\newblock In {\em Proceedings of the IEEE conference on computer vision and
  pattern recognition}, pages 1492--1500, 2017.

\bibitem{metaanchor}
Tong Yang, Xiangyu Zhang, Zeming Li, Wenqiang Zhang, and Jian Sun.
\newblock Metaanchor: Learning to detect objects with customized anchors.
\newblock In {\em Advances in Neural Information Processing Systems}, pages
  320--330, 2018.

\bibitem{reppointsv1}
Ze Yang, Shaohui Liu, Han Hu, Liwei Wang, and Stephen Lin.
\newblock Reppoints: Point set representation for object detection.
\newblock In {\em Proceedings of the IEEE International Conference on Computer
  Vision}, pages 9657--9666, 2019.

\bibitem{unitbox}
Jiahui Yu, Yuning Jiang, Zhangyang Wang, Zhimin Cao, and Thomas Huang.
\newblock Unitbox: An advanced object detection network.
\newblock In {\em Proceedings of the 24th ACM international conference on
  Multimedia}, pages 516--520, 2016.

\bibitem{atss}
Shifeng Zhang, Cheng Chi, Yongqiang Yao, Zhen Lei, and Stan~Z Li.
\newblock Bridging the gap between anchor-based and anchor-free detection via
  adaptive training sample selection.
\newblock In {\em Proceedings of the IEEE/CVF Conference on Computer Vision and
  Pattern Recognition}, pages 9759--9768, 2020.

\bibitem{freeanchor}
Xiaosong Zhang, Fang Wan, Chang Liu, Rongrong Ji, and Qixiang Ye.
\newblock Freeanchor: Learning to match anchors for visual object detection.
\newblock In {\em Advances in Neural Information Processing Systems}, pages
  147--155, 2019.

\bibitem{crossDA}
Yangtao Zheng, Di Huang, Songtao Liu, and Yunhong Wang.
\newblock Cross-domain object detection through coarse-to-fine feature
  adaptation.
\newblock In {\em Proceedings of the IEEE/CVF Conference on Computer Vision and
  Pattern Recognition}, pages 13766--13775, 2020.

\bibitem{objectsaspoints}
Xingyi Zhou, Dequan Wang, and Philipp Kr{\"a}henb{\"u}hl.
\newblock Objects as points.
\newblock {\em arXiv preprint arXiv:1904.07850}, 2019.

\bibitem{autoassign}
Benjin Zhu, Jianfeng Wang, Zhengkai Jiang, Fuhang Zong, Songtao Liu, Zeming Li,
  and Jian Sun.
\newblock Autoassign: Differentiable label assignment for dense object
  detection.
\newblock {\em arXiv preprint arXiv:2007.03496}, 2020.

\bibitem{sapd}
Chenchen Zhu, Fangyi Chen, Zhiqiang Shen, and Marios Savvides.
\newblock Soft anchor-point object detection.
\newblock {\em arXiv preprint arXiv:1911.12448}, 2019.

\bibitem{fsaf}
Chenchen Zhu, Yihui He, and Marios Savvides.
\newblock Feature selective anchor-free module for single-shot object
  detection.
\newblock In {\em Proceedings of the IEEE Conference on Computer Vision and
  Pattern Recognition}, pages 840--849, 2019.

\bibitem{dcnv2}
Xizhou Zhu, Han Hu, Stephen Lin, and Jifeng Dai.
\newblock Deformable convnets v2: More deformable, better results.
\newblock In {\em Proceedings of the IEEE Conference on Computer Vision and
  Pattern Recognition}, pages 9308--9316, 2019.

\end{thebibliography}
}

\clearpage

\appendix

\section{Appendix.}
\subsection{Optimal Transport and Sinkhorn Iteration}\label{sinkhorn}

To ensure the integrity of this paper, we briefly introduce the derivation of the Sinkhorn Iteration algorithm which we emphasize not our contributions and belongs to textbook knowledge.

The mathematical formula of the Optimal Transport problem is defined in Eq.~\ref{origin_formulation}. This is a linear program which can be solved in polynomial time. For dense detectors, however, the resulting linear program is large, involving the square of feature dimensions with anchors in all scales. This issue can be addressed by a fast iterative solution, which converts the optimization target in Eq.~\ref{origin_formulation} into a non-linear but convex form with an entropic regularization term $E$ added:
\begin{alignat}{2}
\begin{split}
\min_{\pi}\quad &\sum\limits_{i=1}^{m}\sum\limits_{j=1}^{n} c_{ij}\pi_{ij} + \gamma E(\pi_{ij}),
\end{split}\label{Eq2}
\end{alignat}
where $E(\pi_{ij})=\pi_{ij} (\log \pi_{ij}-1)$. $\gamma$ is a constant hyper-parameter controlling the intensity of regularization term. According to Lagrange Multiplier Method, the constraint optimization target in Eq.~\ref{Eq2} can be convert to a non-constraint target:
\begin{alignat}{2}
\begin{split}
\min_{\pi}\quad & \sum\limits_{i=1}^{m}\sum\limits_{j=1}^{n} c_{ij}\pi_{ij} + \gamma E(\pi_{ij}) + \\ 
& \alpha_j(\sum\limits_{i=1}^{m} \pi_{ij} - d_j) + \beta_i(\sum\limits_{j=1}^{n} \pi_{ij} - s_i),
\end{split}
\end{alignat}
where $\alpha_j (j=1,2,...n)$ and $\beta_i (i=1,2,...,m)$ are Lagrange multipliers. By letting the derivatives of the optimization target equal to 0, the optimal plan $\pi^{*}$ is resolved as:
\begin{equation}
\begin{split}
\pi_{ij}^{*} = \exp(-{\frac{\alpha_j}{\gamma}})\exp(-{\frac{c_{ij}}{\gamma}})\exp(-{\frac{\beta_i}{\gamma}}).
\end{split}
\end{equation}

Letting $u_j=\exp(-{\frac{\alpha_j}{\gamma}}), v_i=\exp(-{\frac{\beta_i}{\gamma}}), M_{ij}=\exp(-{\frac{c_{ij}}{\gamma}})$, the following constraints can be enforced:
\begin{alignat}{2}
\sum_i \pi_{ij}=u_j(\sum_i M_{ij}v_i)=d_j,\\
\sum_j \pi_{ij}=(u_j\sum_i M_{ij})v_i=s_i.
\end{alignat}

These two equations have to be satisfied simultaneously. One possible solution is to calculate $v_i$ and $u_j$ by repeating the following updating formulas sufficient steps:
\begin{equation}
\begin{split}
u_j^{t+1}=\frac{d_j}{\sum_iM_{ij}v_i^t},\quad v_i^{t+1}=\frac{s_i}{\sum_jM_{ij}u_j^{t+1}}.
\end{split}\label{skiter}
\end{equation}
The updating rule in Eq.~\ref{skiter} is also known as the Sinkhorn-Knopp Iteration. After repeating this iteration $T$ times, the approximate optimal plan $\pi^*$ can be obtained:
\begin{equation}
\begin{split}
\pi^* = diag(v) M diag(u).
\end{split}\label{optimal_assigning_plan}
\end{equation}

$\gamma$ and $T$ are empirically set to 0.1 and 50. Please refer to our code for more details.

\end{document}